\documentclass{article}

\usepackage{PRIMEarxiv}

\usepackage[utf8]{inputenc} 
\usepackage[T1]{fontenc}    
\usepackage{hyperref}       
\usepackage{url}            
\usepackage{booktabs}       
\usepackage{amsfonts}       
\usepackage{nicefrac}       
\usepackage{microtype}      
\usepackage{lipsum}
\usepackage{fancyhdr}       
\usepackage{graphicx}       
\graphicspath{{media/}}     
\usepackage{amssymb}
\usepackage{amsmath}
\usepackage{makecell}
\usepackage{url}
\usepackage{algorithm}
\usepackage{algpseudocode}
\usepackage{afterpage}

\title{Multi-Objective Evolutionary Neural Architecture Search for Recurrent Neural Networks
}

\author{
  Reinhard Booysen \\
  Department of Computer Science \\
  University of Pretoria \\
  Pretoria \\
  \texttt{r.booysen@tuks.co.za} \\
   \And
  Anna Sergeevna Bosman \\
  Department of Computer Science \\
  University of Pretoria \\
  Pretoria \\
  \texttt{annar@cs.up.ac.za} \\
}

\begin{document}
\maketitle

\begin{abstract}
Artificial neural network (NN) architecture design is a nontrivial and time-consuming task that often requires a high level of human expertise. Neural architecture search (NAS) serves to automate the design of NN architectures and has proven to be successful in automatically finding NN architectures that outperform those manually designed by human experts. NN architecture performance can be quantified based on multiple objectives, which include model accuracy and some NN architecture complexity objectives, among others. The majority of modern NAS methods that consider multiple objectives for NN architecture performance evaluation are concerned with automated feed forward NN architecture design, which leaves multi-objective automated recurrent neural network (RNN) architecture design unexplored. RNNs are important for modeling sequential datasets, and prominent within the natural language processing domain. It is often the case in real world implementations of machine learning and NNs that a reasonable trade-off is accepted for marginally reduced model accuracy in favour of lower computational resources demanded by the model. This paper proposes a multi-objective evolutionary algorithm-based RNN architecture search method. The proposed method relies on approximate network morphisms for RNN architecture complexity optimisation during evolution. The results show that the proposed method is capable of finding novel RNN architectures with comparable performance to state-of-the-art manually designed RNN architectures, but with reduced computational demand.
\end{abstract}


\section{Introduction}\label{sec:introduction}
Recurrent neural networks (RNNs) are a set of specialised neural network (NN) architectures that are designed specifically to learn from data with sequential or prominent temporal structures by simulating a discrete-time dynamical system \cite{Mandic2001, Medsker2013, Pascanu2014}. RNNs have been successfully used in solving problems across multiple domains such as forecasting, natural language processing (NLP), load prediction, and more \cite{gpt2, Merity2018, Kong2019, Suzgun2018}.

Designing a NN architecture for a specific problem is a nontrivial task and often requires a high level of human expertise \cite{Wang2020, Yang2019, Zoph2017}. A number of ways have been proposed to automate the task of NN architecture design, collectively referred to as neural architecture search (NAS) methods \cite{Zoph2017, Elsken2018}. NAS aims to automatically find NN architectures for a provided dataset with minimal human intervention, and has already been successful in finding NN architectures that perform comparably to state-of-the-art NN architectures designed by human experts \cite{Elsken2018, Liu2020, Wistuba2019}.

Different approaches to NAS exist for finding well-performing NN architectures, such as reinforcement learning (RL) methods \cite{Zoph2017}, evolutionary algorithm (EA) methods \cite{Lu2020}, gradient-based methods \cite{Chen2020}, and more. Evaluating the performance of a particular NN architecture in NAS is typically based on the corresponding model accuracy \cite{Zoph2017, Chen2020, Klyuchnikov2020}. However, a number of methods have been proposed that consider multiple objectives for evaluating NN architecture performance, which includes NN architecture complexity and the corresponding model computational resource demand \cite{Wang2020, Hu2020, Chu2019}, amongst others. NAS methods that consider multiple objectives for NN architecture performance evaluation typically employ some multi-objective optimisation techniques for finding the best performing NN architectures.

Multi-objective optimisation aims to solve a problem where a solution's effectiveness in reference to a number of objectives determines the solution's quality, where individual objectives may be competing with each other \cite{Smith2015}. In general, for multi-objective problems, assuming the goal is to minimise the respective problems, a vector-valued objective function $F : R^d \rightarrow R^n$ is defined for $n$ objectives, where $n > 1$, and $d$ is the dimension of the decision vector $\mathbf{x}$ \cite{Rudolph1998, Zhou2011, Zitzler2000}. The aim is then to minimise the $\mathbf{y}$ objective vector such that:
\begin{equation*}
    \mathbf{y} = F(\mathbf{x}) = (f_1(\mathbf{x}),...,f_n(\mathbf{x})).
\end{equation*}

When dealing with the optimisation of multiple objectives, the goal is to find the best possible compromise between the objectives \cite{CoelloCoello2002}. Therefore, the advantage of using multi-objective optimisation over single-objective optimisation is that the multi-objective optimisation approach may produce a solution that is more favourable when a reasonable trade-off between the respective objectives is accepted, whereas a single-objective optimisation approach may produce a better quality solution for a single objective.

A multi-objective optimisation approach in NAS aims to optimise multiple objectives that relate to NN architectures, such as the model accuracy, the number of parameters of the model, model inference time, and more. Since multi-objective optimisation techniques aim to find a compromise between the defined objectives, the model accuracy achieved by a multi-objective NAS approach is likely to be worse compared to a NAS approach that considers a single model accuracy objective exclusively. However, NAS approaches that only consider model accuracy objectives disregard the computational resource demand of the models during the search for the optimal NN architecture. Therefore, the trade-off solutions found by the multi-objective NAS approach should produce models with some reasonable model accuracy (potentially worse compared to single-objective approaches), but with reduced computational resource demand.

The practical advantage of a multi-objective NAS approach is that compared to single-objective approaches, models with fewer parameters will be produced if the trade-off between model accuracy and model computational resource demand is acceptable. With RNN architecture search, the majority of the single-objective approaches \cite{Zoph2017, Liu2018, Pham2018} produced models with more than 23M parameters, but none of these approaches were able to produce a model that could outperform a manually designed RNN architecture in terms of model accuracy \cite{Li2019, Yang2018}.

Network morphisms were shown to be a useful tool for evolutionary NAS approaches  \cite{Elsken2018, Cai2018cc}. Generating offspring NN architectures with network morphism is done through the use of network transformation operations, which make structural changes to a cloned parent NN architecture to generate an offspring NN architecture \cite{Cai2018cc, Wei2016a}. These network transformations typically involve the addition of new units to the architecture and the addition of new connections between units in the NN architecture, which are referred to as constructive network transformations \cite{Elsken2018, Cai2018cc}. Destructive network transformation operations, which were introduced by Elsken \textit{et al.} \cite{Elsken2018}, allow for the removal of units and removal of connections between units in the NN architecture, which effectively reduces the overall complexity of the NN architecture.

In this work, we propose a \textbf{M}ulti-\textbf{O}bjective \textbf{E}volutionary algorithm for \textbf{R}ecurrent \textbf{N}eural \textbf{A}rchitecture \textbf{S}earch, dubbed MOE/RNAS, to automatically construct RNN architectures for a provided dataset. The MOE/RNAS algorithm is specifically designed to be capable of optimising some model accuracy-related objectives, along with some RNN architecture complexity-related objectives. The MOE/RNAS algorithm differs from \cite{Bayer2009} in that the MOE/RNAS algorithm is capable of optimising an RNN architecture complexity-related objective with the use of approximate network morphisms. 

Novel contributions of this study are summarised as follows:
\begin{itemize}
    \item MOE/RNAS: a multi-objective EA-based NAS algorithm specifically designed for RNN architecture search is proposed.

    \item Approximate network morphism is implemented to optimise an RNN architecture complexity objective.

    \item A modular RNN architecture block encoding scheme is proposed that is fully capable of catering for destructive RNN network transformations.

    \item An empirical analysis of the MOE/RNAS algorithm's effectiveness to find RNN architectures for three different datasets is conducted.

    \item Experiments show that the proposed MOE/RNAS algorithm is capable of evolving RNN architectures to optimise multiple objectives, which includes at least one RNN architecture complexity objective.

    \item Empirical results show that the proposed MOE/RNAS algorithm can automatically find novel RNN architectures that dominate manually designed RNN architectures when multiple objectives are considered for RNN architecture performance evaluation.  
\end{itemize}

The rest of this paper is structured as follows: Section~\ref{sec:related_work} provides an overview of the relevant background and related work. Section~\ref{sec:method} presents the MOE/RNAS algorithm. Section~\ref{sec:experiments} discusses the experiments and results. Section~\ref{sec:conclusion} concludes the paper. 

\section{Background and Related Work}
\label{sec:related_work}
This section provides an overview of the background and related work and is structured as follows: Section~\ref{sec:background:rnn_architecture} provides a brief discussion of RNN architectures. An overview of EA-based multi-objective optimisation is provided in  Section~\ref{sec:background:emoo}. Section~\ref{sec:background:moo_NAS} discusses the use of multi-objective EAs in NAS. Finally, existing RNN architecture search methods are discussed in Section~\ref{sec:background:RNN_NAS}.

\subsection{Recurrent Neural Network Architecture}
\label{sec:background:rnn_architecture}
Recurrent neural networks (RNNs) are a set of specialised NN architectures that are designed specifically to learn from data with sequential or prominent temporal structures by simulating a discrete-time dynamical system \cite{Mandic2001, Medsker2013, Pascanu2014}. The RNN architecture contains a hidden state component, which serves to provide a feedback connection into the NN. This hidden state allows the RNN to retain information as it progresses through the individual time steps of a particular input sequence \cite{Medsker2013, Pascanu2014}, thereby allowing the RNN to have a form of memory \cite{Chen2016}. 

RNNs face challenges with gradient-based training where input sequences with longer-term dependencies are used. In this case, during backward propagation, the gradient values will either grow exponentially, or go exponentially fast to zero (vanish), such that they become insignificant \cite{Bengio1994, Pascanu2013}. In an attempt to address the RNN's vanishing gradient problem, Hochreiter and Schmidhuber \cite{Hochreiter1997} introduced a novel RNN architecture dubbed Long Short-Term Memory (LSTM). The LSTM deals with the vanishing gradient problem by employing memory cells and gate units \cite{Hochreiter1997}, with the intuition being that the respective units can each form some type of oscillating mechanism, acting like soft switches, to control the amount of information flowing through the network \cite{Kong2019}.

The various gate units of the LSTM are defined by:
\begin{align*}
\mathbf{f}_t &= \sigma(\mathbf{W}_{xf}\mathbf{x}_t + \mathbf{W}_{hf}\mathbf{h}_{t-1} + \mathbf{b}_f), \\
\mathbf{i}_t &= \sigma(\mathbf{W}_{xi}\mathbf{x}_t + \mathbf{W}_{hi}\mathbf{h}_{t-1} + \mathbf{b}_i), \\
\mathbf{o}_t &= \sigma(\mathbf{W}_{xo}\mathbf{x}_t + \mathbf{W}_{ho}\mathbf{h}_{t-1} + \mathbf{b}_o), \\
\mathbf{g}_t &= tanh(\mathbf{W}_{xg}\mathbf{x}_t + \mathbf{W}_{hg}\mathbf{h}_{t-1} + \mathbf{b}_g), \\
\mathbf{c}_t &= \mathbf{f}_t \cdot \mathbf{c}_{t-1} + \mathbf{i}_t \cdot \mathbf{g}_t, \\
\mathbf{h}_t &= \mathbf{o}_t \cdot tanh(\mathbf{c}_t),
\end{align*}
where $\mathbf{f}_t$ is the forget gate, $\mathbf{i}_t$ the input gate, $\mathbf{o}_t$ the output gate, and $\mathbf{g}_t$ is called the input modulation gate. The sigmoid activation function is used for the $\mathbf{f}$, $\mathbf{i}$, and $\mathbf{o}$ gates, which allows the architecture to remain differentiable \cite{Karpathy2015}. $\mathbf{c}_t$ is often referred to as the ``memory cell'' or ``cell state'', and contains information (memory content) from previously encountered inputs of a particular input sequence \cite{Chung2015, Jozefowicz2015}, thereby supplementing the hidden state $h_t$ memory that is implicit in the RNN architecture \cite{Hochreiter1997}.

One notable alternative to the LSTM is the Gated Recurrent Unit (GRU) introduced by Cho \textit{et al.}~\cite{Cho2014} in 2014. The premise of the GRU is that it allows the recurrent unit to capture the dependencies of different time scales \cite{Chung2014}. The GRU employs the same gate-unit philosophy of the LSTM, and the GRU's gate units are defined by:
\begin{align*}
\mathbf{z}_t &= \sigma(\mathbf{W}_{xz}\mathbf{x}_t + \mathbf{W}_{hz}\mathbf{h}_{t-1} + \mathbf{b}_z), \\
\mathbf{r}_t &= \sigma(\mathbf{W}_{xr}\mathbf{x}_t + \mathbf{W}_{hr}\mathbf{h}_{t-1} + \mathbf{b}_r), \\
\mathbf{n}_t &= tanh(\mathbf{W}_{xn}\mathbf{x}_t + \mathbf{W}_{n}(\mathbf{r}_t \cdot \mathbf{h}_{t-1})), \\
\mathbf{h}_t &= \mathbf{z}_t \cdot \mathbf{h}_{t-1} + (1 - \mathbf{z}_t) \cdot \mathbf{n}_t.
\end{align*}
Unlike the LSTM, the GRU does not have a separate memory cell. The GRU uses the update gate $\mathbf{z}_t$ and reset gate $\mathbf{r}_t$ to maintain the unit's memory content, which represents the relevant information from previously encountered input steps of the particular input sequence \cite{Chung2015, Chung2014}. 

\subsection{Evolutionary Multi-Objective Optimisation}
\label{sec:background:emoo}
Multi-objective EAs aim to optimise multiple objectives, which are typically conflicting with each other \cite{engbr2007, Smith2015}. The optimisation of the objectives is done generationally in a survival-of-the-fittest fashion within the population-based paradigm \cite{Ma2019, Olague2016}. Candidate solutions are selected from the population based on their quality, i.e., fitness, and are then combined to produce new offspring solutions for the following generation \cite{Smith2015}.

Deb \textit{et al.} \cite{Deb2002} proposed a fast and elitist nondominated sorting algorithm, called NSGA-II, for evolutionary multi-objective optimisation. The NSGA-II sorts individuals based on their nondomination with respect to the multiple objectives by using a dominance operator \cite{Deb2002}. Dominance states that when one solution dominates another, the dominant solution is at least as good as the other solution for all objectives and additionally, has a strictly better value for at least one of the objectives \cite {Smith2015, Zeng2003}. Additionally, the NSGA-II uses a mechanism called crowding distance assignment to increase the diversity of the population \cite{Deb2002}.

The crowding distance of the NSGA-II represents the distance between candidate solutions, and is used as a density estimator to guide the algorithm towards a uniform population distribution \cite{Deb2002, Chu2018}. After the fittest individuals are selected, offspring are generated by the recombination process of the algorithm with crossover and mutation operators.

\subsection{Multi-Objective Evolutionary NAS}
\label{sec:background:moo_NAS}
Following the introduction of the RL-based NAS method by Zoph and Le \cite{Zoph2017} in 2017, a number of different approaches to NAS have since been proposed, which include EA-based NAS methods \cite{Elsken2018, Lu2020, Lu2020a, Lu2021, Park2020}, amongst others. EA-based NAS methods refer to those NAS methods that employ EAs as their core search space exploration strategy, where the NN architecture search space is traversed in a population-based paradigm \cite{Liu2020}.

The majority of modern EA-based NAS studies focused on convolutional neural network (CNN) architecture search \cite{Elsken2018, Wistuba2019, He2021}. Aside from Bayer \textit{et al.} \cite{Bayer2009}, there have not been any significant investigations into the use of multi-objective EAs for RNN architecture search. Furthermore, most of the methods proposed to make EA-based NAS methods more efficient have not been investigated in the context of RNN architecture search, e.g., network morphism with destructive network transformations \cite{Elsken2018}.

Existing multi-objective EA-based NAS approaches that focus on CNN architecture search is not directly transferable to RNN architecture search, since they lack the ability to represent recurrent connections in the architectures that are explored during the evolutionary search. Additionally, the hidden state of the RNN architecture can not be adequately captured by feedforward NN architecture representations. Therefore, a specialised NAS method is required for representing RNN architectures.

Wei \textit{et al.} \cite{Wei2016a} proposed network morphism as an approach to creating a child network from a parent network such that the function and outputs of the parent NN architecture are preserved in the newly created child network. Therefore, the offspring model's parameters are initialised with the parameters of the corresponding parent model, which have already been optimised during the training of the parent model. The offspring model can then be trained for fewer epochs, making the NN architecture performance evaluation of the NAS method more efficient \cite{Elsken2018, Cai2018cc}. 

Elsken \textit{et al.} \cite{Elsken2018} proposed a Lamarckian Evolutionary algorithm for Multi-Objective Neural Architecture Design (LEMONADE). The LEMONADE algorithm uses a cell-based CNN architecture search space, which is explored through an evolutionary approach that approximates a Pareto front \cite{Elsken2018}. The LEMONADE algorithm does not apply any specific recombination operators such as crossover or mutation, and relies on the concept of network morphism for offspring generation instead \cite{Elsken2018}.

Elsken \textit{et al.} \cite{Elsken2018} noted that previous implementations of network morphism were limited to constructive network transformations, which results in an increased NN architecture complexity. In a multi-objective paradigm where some NN architecture complexity related objectives are considered, \textit{destructive} network transformations are required to optimise NN architecture complexity objective(s).

Elsken \textit{et al.} \cite{Elsken2018} proposed the concept of approximate network morphism to cater for destructive network transformations. Destructive network transformations in the LEMONADE algorithm allow for the removal of units in the architecture and the removal of connections between units \cite{Elsken2018}.

The MOE/RNAS algorithm proposed in this paper is similar to the LEMONADE algorithm, but designed specifically for RNN architecture evolution. Therefore, a novel RNN architecture encoding scheme, as well as a set of network morphisms applicable to RNN architectures, are proposed. Furthermore, the MOE/RNAS algorithm builds on top of the NSGA-II based approach from Bayer \textit{et al.} \cite{Bayer2009} for RNN architecture evolution. Unlike the approach from Bayer \textit{et al.} \cite{Bayer2009}, the MOE/RNAS algorithm also considers RNN architecture complexity objectives, along with appropriate destructive network transformations that allow for the optimisation of RNN architecture complexity-related objectives. 

\subsection{Recurrent Neural Network NAS}
\label{sec:background:RNN_NAS}
Liu \textit{et al.} \cite{Liu2018} proposed a differentiable architecture search method, called DARTS. The DARTS algorithm works by searching for NN architectures in a continuous search space and optimising the NN architectures with respect to their validation set performance by gradient descent \cite{Liu2018}. Liu \textit{et al.} \cite{Liu2018} reported that the DARTS algorithm was able to find CNN and RNN architectures with comparable performance to those found by state-of-the-art NAS methods, but with significantly reduced computational costs. However, the architectures generated by the DARTS algorithm optimise a single model accuracy objective. Therefore, the reduction in computational costs is a by-product as opposed to a design goal, which makes the reduced computational cost potentially unreliable compared to a multi-objective approach, where a reduction in computational cost is one of the objectives considered during optimisation. 

Zoph and Le \cite{Zoph2017} proposed a reinforcement learning (RL) based NAS approach for RNN architecture search wherein they used an RNN controller as the RL agent. The RNN controller explores the RNN cell-based search space by generating a string of computation steps, which includes combination methods and activation functions that are allowed according to the defined search space \cite{Zoph2017}. A cell is then created from the string encoding, which is subsequently used for constructing the RNN architecture \cite{Zoph2017}. Zoph and Le \cite{Zoph2017} defined a cell-based search space for RNN architectures, wherein a single recurrent cell $g$ is described by
\begin{equation}
\label{eq:zoph_search_space}
    \mathbf{h}_t = g_{\theta,\alpha}(\mathbf{x}_t, \mathbf{h}_{t-1}, \mathbf{c}_{t-1}),
\end{equation}
where $\theta$ represents the architecture of the cell $g$, $\alpha$ is the trainable parameters of the architecture, $\mathbf{h}_t$ is the hidden state, $\mathbf{x}_t$ is the input, and $\mathbf{c}_t$ is the cell state at time step $t$. In their study, Zoph and Le \cite{Zoph2017} stacked two recurrent cells to make up the final RNN architecture. Combining multiple inputs to a cell was limited to the use of either addition or elementwise multiplication, and the activation functions were limited to the identity, tanh, sigmoid, and ReLU activation functions \cite{Zoph2017}.

The evaluation of RNN architecture performance considered in \cite{Zoph2017, Liu2018, Klyuchnikov2020} was based on a single objective that relates to the model accuracy. Thus, RNN architecture complexity was not explicitly considered during exploration of the respective RNN architecture search spaces.

Bayer \textit{et al.} \cite{Bayer2009} proposed a method for multi-objective RNN architecture search that was based on a multi-objective EA. However, the multiple objectives considered during their search were based on model accuracy across multiple datasets and did not include any RNN architecture complexity-related objectives, such as the number of parameters that the models have or model inference time \cite{Bayer2009}. Additionally, the particular mutations considered in their approach were limited to constructive network transformations that only allowed for increasing the size of the RNN architectures \cite{Bayer2009}.

Furthermore, a number of existing NAS studies rely on the high-level building blocks of RNN architectures to explore varying connections between existing NN architecture structures \cite{Mo2021}. On the contrary, the method proposed in this work focuses on the optimisation of RNN architectures on a lower level.

To the best of the authors' knowledge, no dedicated studies of a multi-objective EA-based NAS method for novel RNN architecture search exist that also consider an architecture complexity objective. Furthermore, since RNN architecture complexities have not been considered in existing EA-based NAS methods, the use of destructive network transformations has not been studied for RNN architecture evolution.


\section{Multi-Objective Evolutionary algorithm for Recurrent Neural Architecture Search}
\label{sec:method}
\begin{figure*}
    \centering
    \includegraphics[scale=0.5]{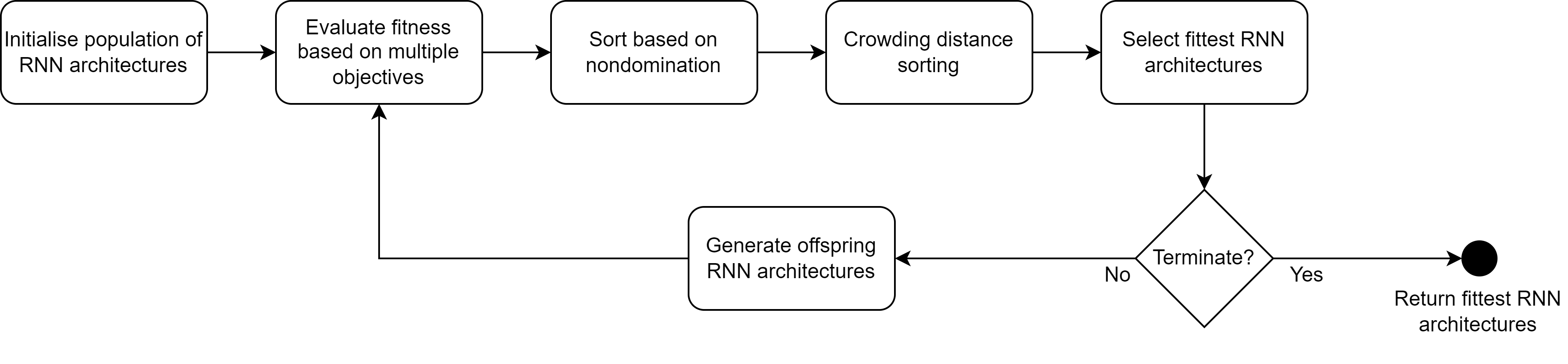}
    \caption{Overview of the MOE/RNAS algorithm}
    \label{fig:moernas_overview}
\end{figure*}

In this section, we present the MOE/RNAS algorithm: a \textbf{M}ulti-\textbf{O}bjective \textbf{E}volutionary algorithm for \textbf{R}ecurrent \textbf{N}eural \textbf{A}rchitecture \textbf{S}earch, to automatically construct RNN architectures for a provided dataset. The MOE/R-NAS algorithm relies on a multi-objective EA that is based on the NSGA-II algorithm for the exploration of the cell-based RNN architecture search space. An overview of the MOE/RNAS algorithm is presented in Figure~\ref{fig:moernas_overview}. The rest of this section is structured as follows: Section~\ref{sec:method:search_space} discusses the search space of the MOE/RNAS algorithm. The search method employed by the MOE/RNAS algorithm for exploration of the search space is discussed in Section~\ref{sec:method:search_method}.

\subsection{Search Space}
\label{sec:method:search_space}
The cell-based RNN architecture search space considered by the MOE/RNAS algorithm draws inspiration from the recurrent cell defined by Zoph and Le \cite{Zoph2017}, as given in Equation~\ref{eq:zoph_search_space} in Section~\ref{sec:background:RNN_NAS}.

The MOE/RNAS algorithm's search space comprises the addition, subtraction, and elementwise multiplication combination methods. The activation functions allowed are the linear, identity, tanh, sigmoid, ReLU, and leaky ReLU activation functions.

The MOE/RNAS algorithm's approach to encoding RNN architectures is discussed below.

\subsubsection{Encoding}

White \textit{et al.} \cite{White2020} identified different categorical encoding schemes that are employed throughout existing NAS methods. The encoding scheme developed for the MOE/RNAS algorithm can be placed within the categorical path encoding scheme that was identified by White \textit{et al.} \cite{White2020}.

When a directed acyclic graph (DAG) representation of the RNN cell is assumed, each node of the DAG can be encoded by a block encoding structure. In the MOE/RNAS algorithm, specifically, an individual block is responsible for performing some operation on one or two inputs. Thus, each block can accept a minimum of one input and a maximum of two inputs. If the block accepts a single input, an activation function must be specified, and the output of the activation function is then used as the output of the particular block. If a block accepts two inputs, a combination method must be specified to indicate how the two inputs must be combined. When a block combines two inputs, the output of the combined inputs can be used as the output of the block, or an optional activation function can be applied to the combined inputs, which is then returned as the output of the block. The MOE/RNAS algorithm's block encoding scheme is illustrated in Figure~\ref{fig:block_encoding}.

An example of a block encoding representation of the basic RNN architecture,
\begin{align*}
    \mathbf{h}_t &= f_h(\mathbf{W}_h\mathbf{x}_t + \mathbf{b}_x + \mathbf{U}_h\mathbf{h}_{t-1} + \mathbf{b}_h),
\end{align*}
with a tanh activation function can be seen in Figure~\ref{fig:basic_rnn_cell}. For the $\mathbf{x}_t$ input to the RNN, an $x_t$ input layer block is created. Similarly, an $h_{t-1}$ input layer block is created for the $\mathbf{h}_{t-1}$ input to the RNN. The $linear\_b$ block that accepts the $x_t$ input layer block as its input, represents the weighted linear activation and bias of the $\mathbf{W}_h\mathbf{x}_t + \mathbf{b}_x$ inputs to the basic RNN architecture. A separate weighted linear activation and bias is used for the $h_{t-1}$ input layer block. The outputs of the two linear activation blocks are then combined using an addition combination block. A $tanh$ activation function is then applied to the output of the combination block, which represents the $f_h$ activation function of the basic RNN architecture. The $\mathbf{h_t}$ output of the RNN architecture at time step $t$ is represented by the $h_{next}$ output layer block, which simply returns the output of the preceding $tanh$ activation block. Note that since the basic RNN architecture does not use the $c_{t-1}$ input layer block, $c_{t-1}$ is simply ignored. 

\afterpage{
    \begin{figure}
        \centering
        \includegraphics[scale=0.35]{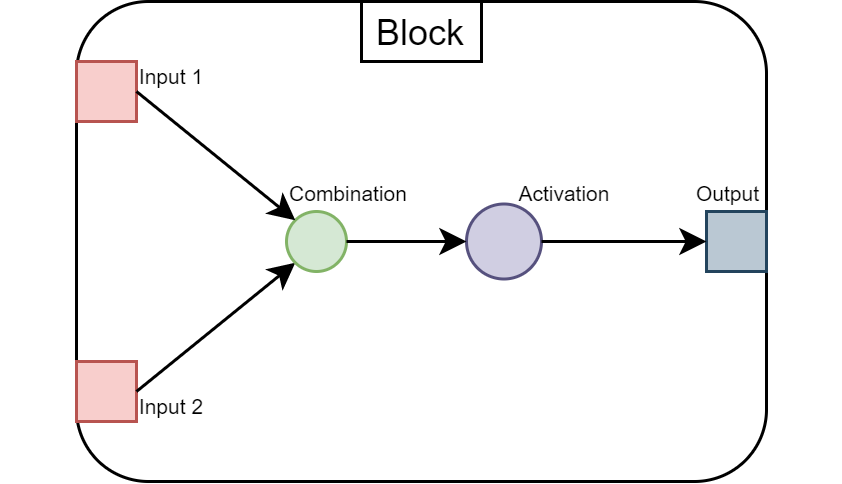}
        \caption{MOE/RNAS algorithm block encoding scheme}
        \label{fig:block_encoding}
    \end{figure}
}


\subsection{Search Method}
\label{sec:method:search_method}
The underlying search strategy that is implemented by the MOE/RNAS algorithm is based on the NSGA-II algorithm. A population of RNN architecture candidate solutions are evolved for a predefined number of generations to find the best performing architecture for the provided dataset, where architecture performance is based on multiple objectives. The MOE/RNAS algorithm maintains a Pareto-optimal front and employs the nondominated rank-based selection operator from the NSGA-II algorithm. Unlike the NSGA-II algorithm, the MOE/RNAS algorithm relies on a network morphism approach to generate offspring as opposed to a multi-parent recombination component.

Whilst similarities may exist between the proposed MOE/RNAS algorithm and genetic programming, it should be noted that the block-based representation method presented in Section~\ref{sec:method:search_space} is only used as a representation for the RNN architecture individuals in the EA's population. During the fitness evaluation stage, an RNN model is constructed based on the blocks of a particular individual; and the blocks themselves have no ability to perform any kind of function, since they merely represent the connections that can exist between the nodes in some RNN architecture. 

The rest of this section discusses the multi-objective EA-based search strategy that is implemented by the MOE/RNAS algorithm. Section~\ref{sec:method:search_method:morphism} discusses the MOE/RNAS algorithm's network morphism approach for generating offspring. The initial population generation procedure is described in Section~\ref{sec:method:search_method:initial}. Section~\ref{sec:method:search_method:fitness} discusses the fitness evaluation of architectures in more detail. The MOE/RNAS algorithm's architecture selection strategy is discussed in Section~\ref{sec:method:search_method:selection}. 

\afterpage{
    \begin{figure}
        \centering
        \includegraphics[scale=0.3]{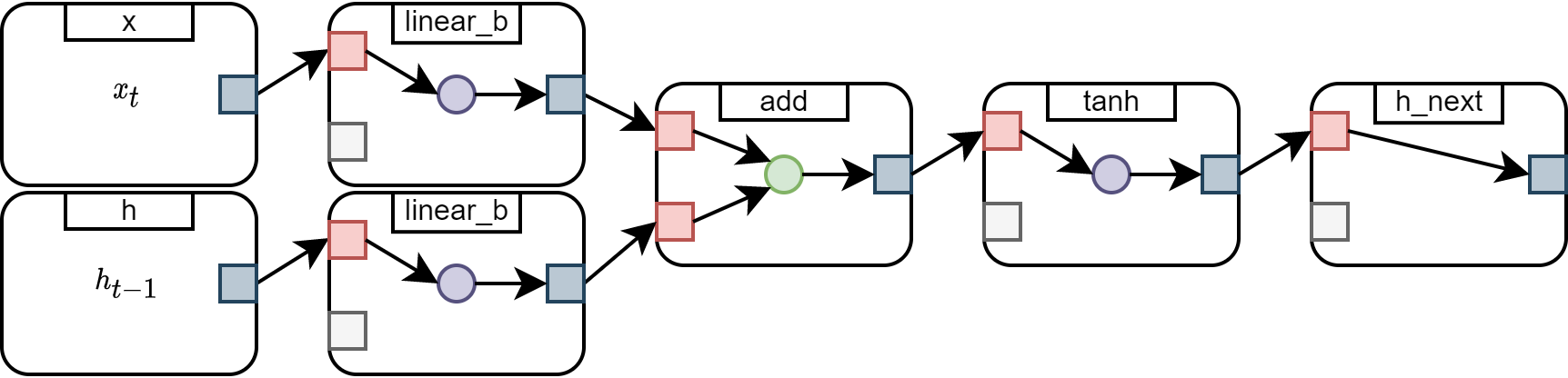}
        \caption{Basic RNN cell block encoding}
        \label{fig:basic_rnn_cell}
    \end{figure}
}


\subsubsection{Approximate Recurrent Neural Network Morphism}
\label{sec:method:search_method:morphism}
The NSGA-II based search space exploration strategy implemented by the MOE/RNAS algorithm employs a network morphism approach instead of a recombination stage with crossover and mutation operators. With network morphism, a single offspring architecture is generated from a single parent architecture, which avoids the complexities associated with performing crossover on multi-parent RNN architectures, as observed in \cite{Bayer2009, Angeline1994}.

Elsken \textit{et al.} \cite{Elsken2018} postulated that the difference in performance between parent and offspring architectures should be low when a maximum number of three network transformations are performed on the offspring architecture. This would allow for a more efficient performance evaluation strategy, since offspring models that share trained parameters with their parent models can be trained for fewer epochs \cite{Elsken2018}.

For each offspring architecture, the MOE/RNAS algorithm randomly selects a number from the range [1, 3], which is then used as the total number of consecutive network transformations that will be applied to that architecture. The network transformations that will be applied are randomly selected with uniform probability. The network transformations implemented by the MOE/RNAS algorithm for offspring generation are described in detail below.

\begin{enumerate}
    \item \textit{add\_unit}: inserts a new activation block between two existing blocks in the architecture. A new block is created and assigned an activation function, which is randomly chosen from: $[linear\_b, linear, identity,\allowbreak sigmoid, tanh, relu, leaky\_relu]$ (see Table~\ref{tab:node-value-descriptions} for descriptions). An existing block $b_r$ is randomly selected from the hidden layer. The newly created block is then inserted between block $b_r$ and one of its inputs; if block $b_r$ has two inputs, one is randomly selected. The effect of this transformation is that an activation will now be applied to selected input from block $b_r$ before the input is passed into block $b_r$. This transformation is an adaptation of the add unit mutation developed by Bayer \textit{et al.} \cite{Bayer2009}. Bayer \textit{et al.} \cite{Bayer2009} restricted the activation function to the linear activation function whereas the MOE/RNAS algorithm randomly selects an activation function from the allowable activation functions listed in Table~\ref{tab:node-value-descriptions}. The output of the newly created block is the result of applying the randomly selected activation to the block's input.

    \item \textit{remove\_unit}: removes a randomly selected activation block from the hidden layer. The \textit{remove\_unit} transformation is effectively the inverse of the \textit{add\_unit} transformation. The single input of the activation block to be removed is set as the input to the subsequently connected blocks that expected the removed block as one of their inputs; this procedure ensures that there are no dangling blocks in the architecture. The \textit{remove\_unit} transformation is a destructive network transformation that allows for the optimisation of the architecture complexity objective.

    \item \textit{add\_connection}: two randomly selected hidden layer blocks are combined. A constraint is enforced to ensure that the two blocks are not already combined or directly connected to each other. A new combination block is then created that accepts both the selected blocks as its inputs; the addition combination method is used for combining the two inputs. All the blocks in the architecture that expect the first of the two randomly selected blocks as their input are identified, and the newly created combination block is set as the replacement input to the identified blocks instead. This transformation is an adaptation of the add connection mutation developed by Bayer \textit{et al.} \cite{Bayer2009}. Bayer \textit{et al.} \cite{Bayer2009} stated that they connected the two units with an identity connection, whereas the MOE/RNAS algorithm introduces the new connection by using the elementwise addition combination method.

    \item \textit{remove\_connection}: removes a randomly selected combination block from the hidden layer; only combination blocks with an addition combination method are considered. When a combination block is removed, it is possible that both of its inputs will be left unused. To deal with this, a procedure is implemented that inspects the architecture to identify the consequences of removing the selected combination block. If it is found that both of the selected combination block's inputs are used by other blocks in the architecture, then the combination block is a good candidate for the \textit{remove\_connection} transformation, and the transformation may therefore proceed without leaving unused blocks in the architecture. If no blocks can be found in the architecture that are good candidates for the \textit{remove\_connection} transformation, then the transformation is simply ignored.

    \item \textit{add\_recurrent\_connection}: introduces a connection between a randomly selected block $b_r$ and either one of the $h_t$ or $c_t$ output layer blocks. This transformation is similar to the \textit{add\_connection} transformation, but aims to specifically add a connection between the randomly selected block and one of the output layer blocks. A newly created combination block with the addition combination method is set as the input to one of the output layer blocks, which is randomly selected. The input from the randomly selected output layer block is assigned as one of the inputs to the newly created combination block. The randomly selected block $b_r$ is then set as the second input to the newly created combination block. This transformation provides for the ability to change an architecture so that it can start using the $c_{t-1}$ input layer block if it has not done so previously.

    \item \textit{change\_activation}: this transformation consists of randomly selecting an activation block from the hidden layer and then simply changing the block's specific activation function to a different activation function, which is randomly selected from the list of allowable activation functions as defined by the search space and summarised in Table~\ref{tab:node-value-descriptions}. The particular block's original activation function is excluded from the list of activation functions to choose from.

    \item \textit{change\_combination}: this transformation consists of randomly selecting a combination block in the hidden layer and then simply changing the block's specific combination method to a different combination method, which is randomly selected from the list of allowable combination methods as defined by the search space and listed in Table~\ref{tab:node-value-descriptions}. The particular block's original combination method is excluded from the list of combination methods to choose from.
\end{enumerate}

\begin{table}[ht]
\begin{center}
\begin{minipage}{\textwidth}
\caption{Descriptions of block values used by the MOE/RNAS block encoding representation}\label{tab:node-value-descriptions}%
\begin{tabular*}{\textwidth}{@{\extracolsep{\fill}}llll@{\extracolsep{\fill}}}
\toprule
Block value & Description & Inputs & Outputs \\
\midrule
x              & The $x_t$ input at step $t$.                                   & 0      & 1       \\
h              & The $h_{t-1}$ input at step $t$.                               & 0      & 1       \\
c              & The $c_{t-1}$ input at step $t$.                               & 0      & 1       \\
h\_next        & The new $h_t$ hidden state for step $t$.                       & 1      & -        \\
c\_next        & The new $c_t$ memory state for step $t$.                       & 1       & -        \\
add            & Addition combination function.                                 & 2      & 1       \\
sub            & Subtraction combination function.                              & 2      & 1       \\
elem\_mul      & \makecell[l]{Elementwise multiplication combination \\ function.}               & 2      & 1       \\
linear\_b      & \makecell[l]{Weighted linear activation function that \\ includes a bias unit.} & 1      & 1       \\
linear         & \makecell[l]{Weighted linear activation function \\ without a bias unit.}       & 1      & 1       \\
identity       & Identity activation function.                                  & 1      & 1       \\
sigmoid        & \makecell[l]{Unipolar sigmoid (logistic) activation \\ function.}               & 1      & 1       \\
tanh           & \makecell[l]{Hyperbolic tangent activation \\ function.}                        & 1      & 1       \\
relu           & \makecell[l]{Rectified Linear Unit (ReLU) activation \\ function.}              & 1      & 1       \\
leaky\_relu    & Leaky ReLU activation function.                                & 1      & 1       \\
$1$ & Integer value.                        &  -      & -        \\ 
\botrule
\end{tabular*}
\end{minipage}
\end{center}
\end{table}

\subsubsection{Initial Population}
\label{sec:method:search_method:initial}
The MOE/RNAS algorithm's procedure for randomly generating an architecture starts with a base RNN architecture, and then performs a number of consecutive network transformations on the architecture. The number of consecutive network transformations that are performed on the architecture is randomly selected from the range $[1,10]$. The base RNN architecture includes the following blocks:
\begin{itemize}
    \item $b_1$, the $x_t$ input layer block;
    \item $b_2$, the $h_{t-1}$ input layer block;
    \item $b_3$, the $c_{t-1}$ input layer block;
    \item $b_4$, a linear activation block that receives $b_1$ as input;
    \item $b_5$, a linear activation block that receives $b_2$ as input;
    \item $b_6$, a linear activation block that receives $b_3$ as input;
    \item $b_7$, a block that receives blocks $b_4$ and $b_5$ as inputs and combines these inputs, the combination function is randomly chosen from $[add, sub,\allowbreak elem\_mul]$ (see Table~\ref{tab:node-value-descriptions});
    \item $b_8$, an activation block that receives $b_7$ as input, the activation function is randomly chosen from $[linear\_b,\allowbreak linear, identity, sigmoid, tanh, relu,\allowbreak leaky\_relu]$ (see Table~\ref{tab:node-value-descriptions});
    \item $b_9$, the $h_t$ output layer block that receives $b_8$ as input;
    \item $b_{10}$, the $c_t$ output layer block that receives $b_6$ as input.
\end{itemize}
The \textit{remove\_unit} and  \textit{remove\_connection} network transformations are excluded when randomly generating architectures for the initial population. This is done so that only constructive network transformations are allowed, which will effectively promote a more diverse initial population.

Each architecture in the population is assigned a unique identifier. The unique identifier is generated using the template $X\_c$, where $X$ is a short string that is assigned to the initial architecture, and $c$ is an integer to represent the count of the particular architecture. The initial architectures will start with a $c$ value of $0$, and subsequently generated offspring architectures will have increased values for $c$. Randomly generated architectures are assigned an $X$ value of $rdmY$, where $Y$ represents a unique integer assigned to that particular architecture. Therefore, the first randomly generated architecture in the initial population will be assigned the identifier $rdm0\_0$, the second randomly generated architecture in the initial population $rdm1\_0$, and so on. If existing architectures are supplied to be included in the initial population, they will be assigned appropriate identifiers. For example, if the LSTM architecture is supplied to be included in the initial population, the architecture's identifier will be $LSTM\_0$.

After the initial population generation procedure has concluded, the fitness values for each of the individual architectures are evaluated based on the provided objectives. The MOE/RNAS algorithm's fitness evaluation process is described in detail in the following section. 


\subsubsection{Fitness Evaluation}
\label{sec:method:search_method:fitness}
The fitness evaluation procedure implemented by the MOE/RNAS algorithm assumes the responsibility of the NAS performance estimation component. Thus, the performances of the architectures are based on the fitness values calculated by the MOE/RNAS algorithm's EA fitness evaluation method.

The fitness of architectures in the population is calculated based on the objectives provided. It is expected for one of the objectives to represent the architecture's achieved accuracy after being trained and validated on relevant subsets of the provided dataset. Furthermore, at least one objective should be included that relates to architecture complexity. The MOE/RNAS algorithm supports the following architecture complexity related objectives:
\begin{itemize}
    \item the number of blocks that the architecture contains;
    \item the number of parameters of the model;
    \item the model inference time, i.e., how long the model takes for a forward propagation of a single input pattern.
\end{itemize}

The MOE/RNAS algorithm does not implement any specific techniques that predict model accuracy. Instead, the MOE/RNAS algorithm relies on a parameter sharing technique, where offspring models are initialised with the parameters of their respective parent models. As a result of the network morphism approach for generating offspring architectures along with parameter sharing between parents and offspring, the offspring models can be trained for fewer epochs. The performance difference between parents and offspring is based on the observations reported by Elsken \textit{et al.} \cite{Elsken2018}, when a network morphism approach is used for generating offspring.

The MOE/RNAS algorithm performs the selection of architectures based on their respective fitness values and ranking during the evolutionary cycle, which is done according to the selection operators of the NSGA-II algorithm. The next section provides an overview of how architecture selection is performed by the MOE/RNAS algorithm.


\subsubsection{Selection}
\label{sec:method:search_method:selection}
After the fitness values for each of the individuals in the population have been evaluated, the individuals are sorted based on their nondomination and placed into appropriate Pareto fronts. The nondominated sorting of individuals in the population based on their objective values is done according to the NSGA-II nondominated sorting method, without any adaptation.

Survivor selection is performed in the same way as it is done by the NSGA-II algorithm. The NSGA-II algorithm generates $N$ offspring, which results in a $2N$ sized combined population from which survivor selection is performed. With larger values of $N$,  a significant number of models need to be trained and validated. The MOE/RNAS algorithm has an input parameter that can be used to specify the maximum number of parents to select for offspring generation. The top performing architectures are selected as parents if the aforementioned input parameter is smaller than $N$. 

Pseudocode for the MOE/RNAS algorithm is given in Algorithm~\ref{alg:nas_approach}.
\begin{algorithm}
\caption{MOE/RNAS Algorithm}\label{alg:nas_approach}
\begin{algorithmic}
\State Inputs: $N$, $\phi$, termination condition, objectives, seeds;
\State $i \Leftarrow 0$; \Comment{initialise generation counter}
\State $\Upsilon \Leftarrow \emptyset$; \Comment{initialise empty archive for keeping track of previously evaluated architectures}
\State $P \Leftarrow initialisePopulation(N, seeds)$ \Comment{initialise parent population of size N, include seed architectures} 
\State $Q \Leftarrow$ initialise offspring population to $\emptyset$;

\State $f \Leftarrow evaluateFitness(P)$; 
\State $[F_1, F_2, ...] \Leftarrow nondominatedSort(f, \hat{f}(P))$ \Comment{calculate and construct Pareto-fronts based on nondomination}
\State $\Upsilon \Leftarrow \Upsilon \cup P$;

\State $p \Leftarrow tournamentSelection(P, [F_1, F_2, ...])$
\State $Q \Leftarrow generateOffspring(p, \phi)$ \Comment{generate $\phi$ number of offspring architectures}

\While{termination condition not met}
\State $f' \Leftarrow evaluateFitness(Q)$;
\State $[F_1, F_2, ...] \Leftarrow nondominatedSort(f \cup f' , \hat{f}(P) \cup \hat{f}(Q))$

\State $dist \Leftarrow crowdingDistanceAssignment(F_1, F_2, ...)$
\State $P \Leftarrow survivorSelection(P \cup Q, [F_1, F_2, ...], dist, N)$

\State $i \Leftarrow i + 1$; \Comment{update generation counter}
\State $\Upsilon \Leftarrow \Upsilon \cup Q$;

\State $Q \Leftarrow generateOffspring(P, \phi)$
\EndWhile
\end{algorithmic}
\end{algorithm}


\section{Empirical Results}
\label{sec:experiments}
This section presents the results of the MOE/RNAS algorithm's ability to find and evolve novel RNN architectures. The MOE/RNAS algorithm was set to optimise the following two objectives: (i) the model accuracy objective, and (ii) an RNN architecture complexity objective, which was based on the number of blocks that the architecture contained. The following tasks were considered:
\begin{enumerate}
    \item A standard word-level NLP task based on the Penn Treebank dataset. The Penn Treebank dataset is often used as a benchmark in RNN NAS research \cite{Zoph2017, Klyuchnikov2020, Li2019, Liu2018}. Although it is unlikely for any current NAS method to find a novel RNN architecture that outperforms state-of-the-art RNN architectures that were designed by human experts \cite{Klyuchnikov2020, He2021}, an EA-based RNN architecture search method has not been implemented on the Penn Treebank dataset.

    The Penn Treebank dataset contains 10 000 unique words, and is therefore a good candidate for testing whether the RNN architectures evolved by the MOE/RNAS algorithm can learn from the provided dataset. Since the models are expected to predict the next word in the sequence, model accuracy highly depends on what the model has learned from the data during training. 
    
    \item A sequence learning task based on artificially generated strings from a context-sensitive language, which was previously used in the study by Bayer \textit{et al.} \cite{Bayer2009}. The training and testing datasets consist of strings that are generated from the $a^nb^nc^n$ context-sensitive language, where the value of $n$ is randomly selected from the range $\{1..10\}$ for each string.

    By artificially generating the sequence learning task's dataset from a context-sensitive language, the MOE/RNAS algorithm is inadvertently presented with a challenge to evolve RNN architectures with sufficient memory capabilities, such that they can learn the significance of the determinism of the particular context-sensitive language. Therefore, this dataset is useful for gaining a better understanding of the relationship between multi-objective RNN architecture evolution and model accuracy. 

    \item A sentiment analysis task that is based on the ACL-IMBD \cite{Maas2011} dataset. This dataset contains 50 000 sentences, each of which has either a positive or a negative sentiment.
    
\end{enumerate}

For all three tasks, the RNN architecture complexity objective was based on the number of blocks that an architecture comprised. The model accuracy objective that was used is discussed in more detail under the relevant sections below.

Technical implementation details for the experiments were as follows:
\begin{enumerate}
    \item All the source code implementations\footnote{Source code implementation of the MOE/RNAS algorithm is available at \url{https://github.com/reinn-cs/rnn-nas}.} of this study were developed using the Python programming language and the PyTorch \cite{Paszke2019} framework. 
    \item Experiments were run on a system with a single Nvidia V100 16GB GPU.
\end{enumerate}

The rest of this section is structured as follows: Section~\ref{sec:experiments:ptb} discusses the results for the word-level NLP task that is based on the Penn Treebank dataset. The sequence learning task results are discussed in Section~\ref{sec:experiments:cfl}. Section~\ref{sec:experiments:sentiment} discusses the results for the sentiment analysis task. The observations from the experiments are summarised in Section~\ref{sec:experiments:discussion}.

\subsection{Word-Level Natural Language Processing Task}
\label{sec:experiments:ptb}
A total of three experimental runs were performed for the word-level NLP task. One experimental run included the LSTM and GRU architectures in the initial population. Two experimental runs were performed where the LSTM and GRU architectures were excluded from the initial population. The limited number of experimental runs were due to the inherently high computational resource demand of NAS.

The performance of a model implemented for the standard word-level NLP task is calculated based on how well the model is able to predict the next word, which is commonly represented by a metric called \textit{perplexity} \cite{Jozefowicz2016, Jurafsky2014}. Perplexity measures how accurately a model can predict the next word, such that for a given test set $D_G = d_1d_2...d_Q$, the perplexity is calculated by:

\begin{align*}
PP(D_G) = P(d_1d_2...d_Q)^{-\frac{1}{Q}} 
 = \sqrt[Q]{\frac{1}{P(d_1d_2...d_Q)}}
\end{align*}
normalised by the number of words \cite{Jurafsky2014}. As noted by Jurafsky and Martin \cite{Jurafsky2014}, the chain rule can be used to expand the probability of $D_G$ such that:

\begin{equation*}
    PP(D_G) = \sqrt[Q]{\prod^Q_{i=1} \frac{1}{P(d_i \mid d_1...d_{i-1})}}.
\end{equation*}
Therefore, the model accuracy objective considered during this experiment is based on the perplexities achieved by the respective models on the Penn Treebank dataset.

The RNN architectures created by the MOE/RNAS algorithm were not stacked (i.e. repeated) during model creation, and instead, each model contained a single instance of the corresponding RNN architecture, i.e., a single cell. The models were implemented with an embedding layer dimension of 650 and a hidden layer dimension of 650, which was adopted from \cite{Nugaliyadde2019}. A batch size of 20 was used during model training, and the RNN models were unrolled for 35 time steps during backpropagation training. For each model, a dropout layer was included to randomly zero some of the elements of the input with a probability of 0.5. Models were trained for 30 epochs using a stochastic gradient descent training method. Training of the models started with a learning rate of 20, and the learning rate was reduced when the model performance started stagnating; the specific learning rate reduction was adopted from \cite{Merity2018}.  Offspring model parameters were initialised with their respective parent model parameters. If the initial test perplexity difference between an offspring model and its parent was more than 5 perplexity points, the offspring model was trained for 30 epochs. Alternatively, offspring models were only trained for 5 epochs.

The initial population included the basic RNN, LSTM, and GRU architectures. 97 RNN architectures were uniformly sampled from the search space, which resulted in a total population size of 100. The search was terminated after 30 generations and the total search cost was 8.25 GPU days for one experimental run.

\begin{figure}
    \centering
    \includegraphics[height=65mm,keepaspectratio]{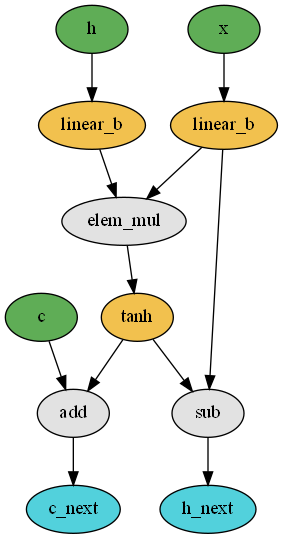}
    \caption{The rdm68\_45 architecture evolved by the MOE/RNAS algorithm}
    \label{fig:rdm68_45_arch}
\end{figure}

The Pareto-optimal RNN architectures found by the MOE/RNAS algorithm are listed in Table~\ref{tab:chpc_12_28_one}. The rdm68\_45 architecture achieved the best test perplexity of 92.704 across all architectures that were generated and evolved by the MOE/RNAS algorithm; the rdm68\_45 architecture is illustrated in Figure~\ref{fig:rdm68_45_arch} (refer to Section~\ref{sec:method:search_method:initial} for RNN architecture identifier notation).

The LSTM outperformed the rdm68\_45 architecture by 8.76 perplexity points, however, the rdm68\_45 architecture has 14 blocks less compared to the LSTM. Furthermore, the rdm68\_45 architecture has 2.5M fewer parameters compared to the LSTM, which makes the rdm68\_45 architecture significantly more efficient compared to the LSTM. The reduced computational demand of the rdm68\_45 architecture justifies the reasonable 8.76 perplexity point trade-off compared to the better performing LSTM architecture.

\begin{figure}
    \centering
    \includegraphics[height=35mm,keepaspectratio]{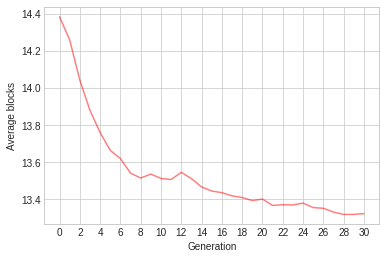}
    \caption{Word-level NLP task average number of blocks per generation for a single run}
    \label{fig:chpc_12_28_one:avg_blocks}
\end{figure}

\begin{figure}
    \centering
    \includegraphics[height=35mm,keepaspectratio]{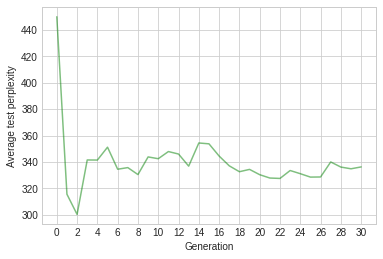}
    \caption{Word-level NLP task average test perplexity per generation for a single run}
    \label{fig:chpc_12_28_one:avg_ppl}
\end{figure}

\begin{figure}
    \centering
    \includegraphics[height=35mm,keepaspectratio]{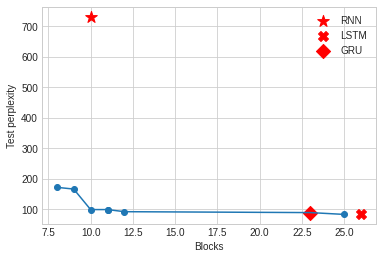}
    \caption{The Pareto-front of the word-level NLP task experiment. RNN and LSTM architectures are included for reference}
    \label{fig:chpc_12_28_one_pareto}
\end{figure}

The results show that the MOE/RNAS algorithm succeeded in optimising the architecture complexity objective by maintaining a consistent decrease in the average number of blocks across the population of architectures per generation, which can be seen in Figure~\ref{fig:chpc_12_28_one:avg_blocks}. The average test perplexity per generation did not exhibit a similar trend, but neither did it worsen across the generations, as illustrated in Figure~\ref{fig:chpc_12_28_one:avg_ppl}. Therefore, the MOE/RNAS algorithm was able to optimise the architecture complexity objective without negatively influencing the model accuracy objective across the 30 generations. The best performing RNN architectures that were evolved by the MOE/RNAS algorithm dominated the manually designed LSTM architecture in terms of Pareto-optimality, while the GRU architecture remained non-dominated across the 30 generations; the final Pareto-front for this experiment can be seen in Figure~\ref{fig:chpc_12_28_one_pareto}.

\begin{figure}
    \centering
    \includegraphics[width=0.45\textwidth]{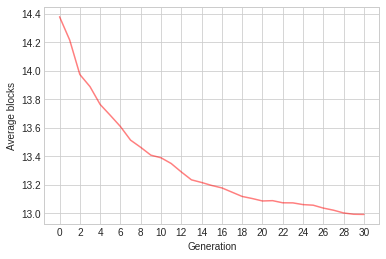}
    \caption{Average number of blocks per generation for control experiment}
    \label{fig:chpc_chpc_12_28_two:avg_blocks}
\end{figure}

\textbf{Control Experiment - Exclude LSTM and GRU From Initial Population:} In this experiment, the LSTM and GRU architectures were not included in the initial population. Therefore, the initial population comprised the basic RNN architecture and 99 randomly generated architectures. The results show that the MOE/RNAS algorithm was able to consistently optimise the RNN architecture complexity objective as the EA progressed, which can be seen in Figure~\ref{fig:chpc_chpc_12_28_two:avg_blocks}. Despite the average test perplexity per generation exhibiting some improvement as the EA progressed, the average test perplexity per generation is higher compared to the average test perplexity per generation from the previous experiment. From the Pareto-optimal architectures listed in Table~\ref{tab:chpc_12_28_two}, it can be seen that the best performing RNN architecture evolved during this experiment achieved a test perplexity of 94.318, which is 1.6 perplexity points worse compared to the best performing architecture evolved by the MOE/RNAS algorithm in the previous experiment.

After 30 generations, the total search cost of this experiment was 6.25 GPU days, which is better compared to the 8.25 GPU days search cost of the previous experiment. By including the LSTM and GRU architectures in the previous experiment's initial population, a number of offspring architectures were generated from the LSTM and GRU architectures, which contributed towards longer model training times and inevitably led to a higher search cost.

This control experiment was then repeated with the same configuration. The best architecture found during this experimental run was generated after 20 generations and achieved a test perplexity of 91.304. From the Pareto-optimal architectures listed in Table~\ref{tab:chpc_12_28_two_second}, it is observed that the MOE/RNAS algorithm was able to consistently optimise the complexity objective, with 14 blocks being the highest number of blocks amongst the Pareto-optimal architectures.

\begin{table}[ht]
\begin{center}
\begin{minipage}{250pt}
\caption{The word-level NLP task's Pareto-optimal architecture performances}\label{tab:chpc_12_28_one}%
\begin{tabular*}{250pt}{@{\extracolsep{\fill}}lcc@{\extracolsep{\fill}}}
\toprule
Architecture & Test perplexity & Number of blocks \\
\midrule
LSTM\_58     & \textbf{83.782} & 25      \\
LSTM\_0      & 83.945          & 26      \\
GRU\_0       & 89.766          & 23      \\
rdm68\_45    & 92.704          & 11      \\
rdm8\_0      & 99.625          & 10      \\
rdm8\_3      & 169.047         & 9       \\
rdm8\_190    & 172.487         & \textbf{8} \\
\botrule
\end{tabular*}
\end{minipage}
\end{center}
\end{table}

\begin{table}[ht]
\begin{center}
\begin{minipage}{250pt}
\caption{The word-level NLP task's control experiment Pareto-optimal architecture performances}\label{tab:chpc_12_28_two}%
\begin{tabular*}{250pt}{@{\extracolsep{\fill}}lcc@{\extracolsep{\fill}}}
\toprule
Architecture & Test perplexity & Number of blocks \\
\midrule
rdm35\_108   & \textbf{94.318} & 14     \\
rdm35\_116   & 96.588          & 15     \\
rdm21\_62    & 97.318          & 19     \\
rdm21\_40    & 99.476          & 17     \\
rdm5\_0      & 101.313         & 10     \\
BASIC\_18    & 165.862         & 9      \\
rdm28\_26    & 170.537         & \textbf{8} \\
\botrule
\end{tabular*}
\end{minipage}
\end{center}
\end{table}

\begin{figure}
    \centering
    \includegraphics[width=0.45\textwidth]{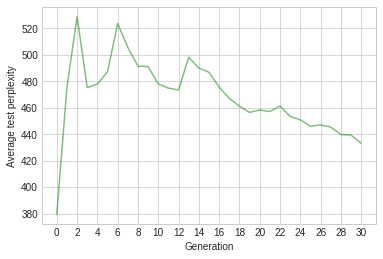}
    \caption{Average test perplexity per generation for control experiment}
    \label{fig:chpc_chpc_12_28_two:avg_ppl}
\end{figure}

\begin{table}[ht]
\begin{center}
\begin{minipage}{250pt}
\caption{The second run of the word-level NLP task's control experiment Pareto-optimal architecture performances}\label{tab:chpc_12_28_two_second}%
\begin{tabular*}{250pt}{@{\extracolsep{\fill}}lcc@{\extracolsep{\fill}}}
\toprule
Architecture & Test perplexity & Number of blocks \\
\midrule
rdm6\_104   & \textbf{91.304}   & 14     \\
rdm6\_23    & 91.379            & 13     \\
rdm6\_90    & 92.088            & 12     \\
rdm6\_76    & 93.582            & 11     \\
rdm46\_388  & 101.820           & 10     \\
rdm46\_86   & 103.486           & 9      \\
rdm79\_79   & 167.251           & 8 \\
rdm46\_25   & 171.414           & \textbf{7} \\
\botrule
\end{tabular*}
\end{minipage}
\end{center}
\end{table}

\subsection{Sequence Learning Task Based on Artificially Generated Data}
\label{sec:experiments:cfl}
This section discusses the experimental results obtained after implementing the MOE/RNAS algorithm to search for and optimise RNN architectures for a sequence learning task. The dataset used for this task was generated from the $a^nb^nc^n$ context-sensitive language, which is the same context-sensitive language used by Bayer \textit{et al.} \cite{Bayer2009} in their multi-objective EA-based RNN architecture search method.

The training and testing datasets consisted of strings that were generated from the $a^nb^nc^n$ context-sensitive language. The training dataset consisted of 500 strings generated from the language $a^nb^nc^n$, where the value of $n$ was randomly selected from the range $\{1..10\}$ for each string. The testing dataset was limited to 100 strings, and the values for $n$ were randomly chosen from the range $\{1..10\}$. For example, $n = 3$ results in the string $aaabbbccc$ being generated. One single input sequence from either the training or testing datasets consisted of a string where each character of that particular string was considered an input in the input sequence. For each of the input sequences, the model was presented with an arbitrary sub-string of the particular input sequence, and the model was then expected to predict the remaining characters of the string from that particular input sequence.

The model accuracy objective considered throughout this experiment was based on the mean squared error (MSE) loss obtained by the model on the generated testing dataset, after the model was trained on the training dataset. In this experiment, the RNN architectures created by the MOE/RNAS algorithm were not stacked, and each model contained a single instance of the corresponding RNN architecture. The models were implemented with a hidden layer dimension of 128, and since the dataset is relatively small, batching was not implemented during training. The models were unrolled for the full length of the input sequence, which was up to a maximum of 10 steps. Training of the models was done using the backpropagation through time training algorithm with a stochastic gradient descent optimisation technique and a learning rate of 0.01.

Parent selection was limited to the top 25 architectures of the Pareto front (see Section~\ref{sec:method:search_method:selection}). Thus, only 25 offspring architectures were produced for each generation. During offspring generation, up to ten consecutive network transformations were allowed per architecture (see Section~\ref{sec:method:search_method:morphism}). Therefore, fewer offspring architectures were generated and a higher number of consecutive network transformations were allowed compared to the previous word-level NLP experiments. This was done specifically to gain some insight into the multi-objective RNN morphism approach employed by the MOE/RNAS algorithm.

The initial population included the basic RNN, LSTM, and GRU architectures. 97 RNN architectures were uniformly sampled from the search space, which resulted in a total population size of 100. The search was terminated after 15 generations, which resulted in a total search cost of 0.33 GPU days.

\begin{figure}
    \centering
    \includegraphics[width=0.45\textwidth]{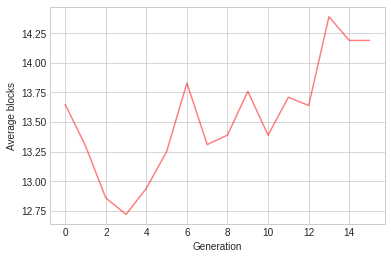}
    \caption{Average number of blocks per generation observed for the sequence learning task}
    \label{fig:chpc_11_03_cfl:avg_blocks}
\end{figure}

According to Figure~\ref{fig:chpc_11_03_cfl:avg_blocks}, the MOE/RNAS algorithm struggled to maintain an optimised RNN architecture complexity objective across the population of RNN architectures, since the average number of blocks per generation increased as the evolutionary cycle progressed. This was a result of the increased number of consecutive network transformations allowed during network morphism.

\begin{figure}
    \centering
    \includegraphics[width=0.45\textwidth]{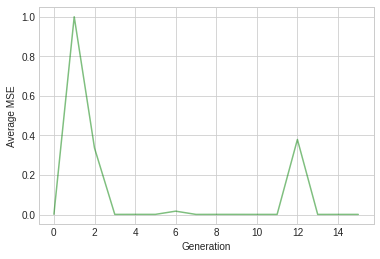}
    \caption{Average MSE loss per generation observed for the sequence learning task}
    \label{fig:chpc_11_03_cfl:avg_mse}
\end{figure}

Although the average MSE per generation shown in Figure~\ref{fig:chpc_11_03_cfl:avg_mse} does not exhibit a noticeable trend, the MOE/RNAS algorithm was able to successfully optimise the model accuracy objective. According to the performances of the Pareto-optimal architectures listed in Table~\ref{tab:cfl_run_1}, it is observed that the MOE/RNAS algorithm was able to find and evolve a novel RNN architecture that outperformed the LSTM in both the model accuracy and architecture complexity objectives.

\begin{figure}
    \centering
    \includegraphics[scale=0.3]{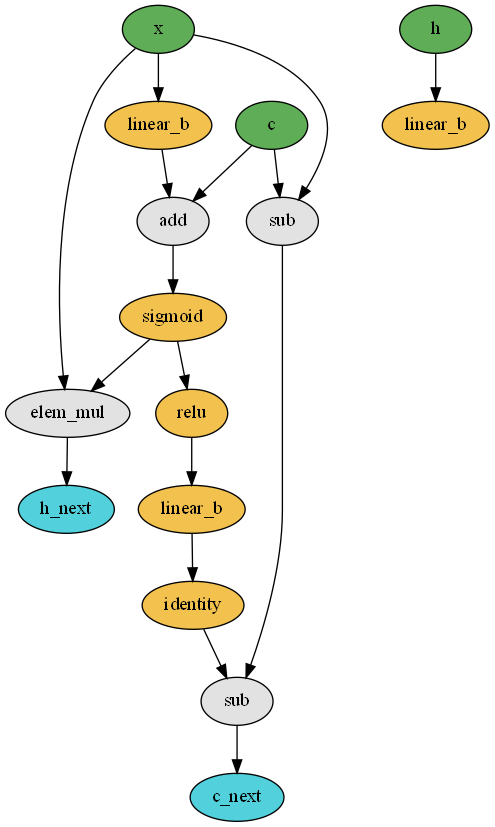}
    \caption{The rdm82\_21 architecture evolved by the MOE/RNAS algorithm}
    \label{fig:rdm82_21}
\end{figure}

The rdm82\_21 architecture shown in Figure~\ref{fig:rdm82_21} is particularly interesting. During the network morphism, the validity of an architecture is determined based on its use of the hidden state blocks, as previously discussed in Section~\ref{sec:method:search_method:morphism}. There is no verification performed to verify that a path exists exactly from the $h_{t-1}$ input layer block to the $h_t$ output layer block. The evolutionary algorithm exploited this during the evolution of the architecture rdm82\_21. The $h_t$ output layer block has at least one input, and there is at least one other block that uses the $h_{t-1}$ block as its input. Thus, the generation of this particular architecture did not violate any of the predefined constraints.

The interesting observation from the rdm82\_21 architecture is that it still maintains a recursive structure through the path of the $c_{t-1}$ input layer block, which eventually reaches the $h_t$ output layer block. The output of the $h_t$ output layer block at the last input of the input sequence is used as the output of the architecture. Therefore, the architecture effectively used the $c_t$ output layer block as a substitute for the hidden state.

\begin{figure}
    \centering
    \includegraphics[width=0.45\textwidth]{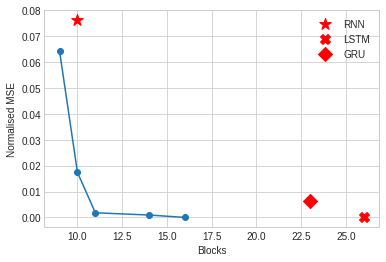}
    \caption{Pareto-front of the sequence learning task, which also includes the RNN, GRU, and LSTM architectures for reference}
    \label{fig:chpc_11_03_pareto}
\end{figure}

Thus, despite being unable to optimise the average number of blocks per generation, the MOE/RNAS algorithm was able to find and evolve novel RNN architectures that dominated the RNN, GRU, and LSTM architectures in terms of Pareto-optimality in less than 15 generations; the final Pareto-front for this experiment can be seen in Figure~\ref{fig:chpc_11_03_pareto}.

\begin{table}[ht]
\begin{center}
\begin{minipage}{250pt}
\caption{Pareto-optimal architecture performances of the sequence learning task, which includes the performance of the LSTM architecture for reference}\label{tab:cfl_run_1}%
\begin{tabular*}{250pt}{@{\extracolsep{\fill}}lcc@{\extracolsep{\fill}}}
\toprule
Architecture & MSE loss & Number of blocks \\
\midrule
rdm82\_21 & \textbf{0.000014} & 15  \\
LSTM\_0      & 0.000157 & 26           \\
rdm82\_18 & 0.000939 & 14           \\
rdm1\_21  & 0.001806 & 11           \\
rdm43\_0  & 0.017431 & 10           \\
BASIC\_29 & 0.064309 & \textbf{9}  \\
\botrule
\end{tabular*}
\end{minipage}
\end{center}
\end{table}

\textbf{Control Experiment - Reduced Number of Consecutive Network Transformations Allowed:} In this experiment, a maximum number of three consecutive transformations were considered during network morphism. Additionally, 100 offspring architectures were created for each generation.

\begin{figure}
    \centering
    \includegraphics[width=0.45\textwidth]{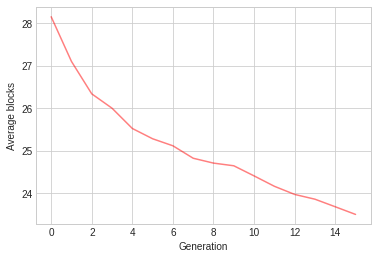}
    \caption{Average number of blocks per generation for the sequence learning control experiment}
    \label{fig:chpc_01_12_cfl:avg_blocks}
\end{figure}

\begin{figure}
    \centering
    \includegraphics[width=0.45\textwidth]{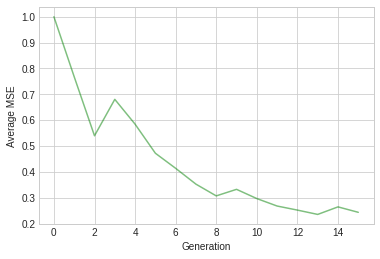}
    \caption{Average MSE loss per generation for the sequence learning control experiment}
    \label{fig:chpc_01_12_cfl:avg_mse}
\end{figure}

Figure~\ref{fig:chpc_01_12_cfl:avg_blocks} and Figure~\ref{fig:chpc_01_12_cfl:avg_mse} show the favourable trends in terms of the average number of blocks per generation and the average MSE per generation across the 15 generations, respectively. Furthermore, the MOE/RNAS algorithm was able to maintain a consistent decrease in the average number of blocks per generation while simultaneously optimising the model accuracy objective. Thus, the number of consecutive network transformations considered during network morphism has a clear contribution towards the multi-objective optimisation of the RNN architectures.

Table~\ref{tab:cfl-01-12-1-pareto} lists the Pareto-optimal architecture performances. Apart from the BASIC\_0 architecture, all other architectures in the Pareto front listed in Table~\ref{tab:cfl-01-12-1-pareto} are offspring architectures that were optimised from the randomly generated architectures during the initialisation of the population.

\begin{figure}
    \centering
    \includegraphics[width=0.45\textwidth]{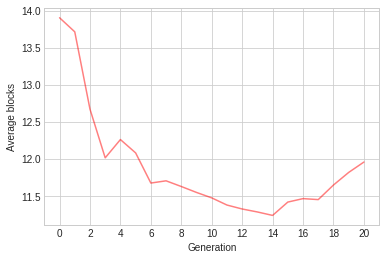}
    \caption{Average number of blocks per generation for the second run of the sequence learning control experiment}
    \label{chpc_cfl_2_blocks}
\end{figure}

\begin{figure}
    \centering
    \includegraphics[width=0.45\textwidth]{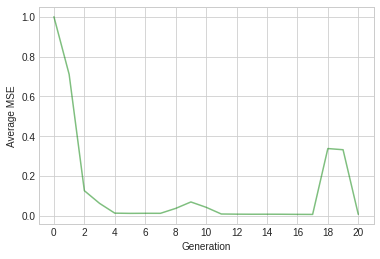}
    \caption{Average MSE loss per generation for second run of the sequence learning control experiment}
    \label{chpc_cfl_2_mse}
\end{figure}

The sequence learning task was repeated with the same configuration, but for 20 generations as opposed to 15 generations. From the results of this run it was observed that the average number of blocks per generation appeared to have reached an optimum after 14 generations. From the 15th generation onwards, the average number of blocks per generation started increasing, which can be seen in Figure~\ref{chpc_cfl_2_blocks}. Although the second run of the control experiment did not exhibit a similar trend in terms of the model accuracy objective to that of the initial run of the control experiment, the MOE/RNAS algorithm was still able to maintain a relatively low average MSE per generation throughout the run, which can be seen in Figure~\ref{chpc_cfl_2_mse}.

Therefore, the MOE/RNAS algorithm is clearly capable of generationally optimising multiple RNN architecture objectives when the appropriate configuration is considered, such as the number of consecutive network transformations allowed during network morphism.

\begin{table}[ht]
\begin{center}
\begin{minipage}{250pt}
\caption{Pareto-optimal architecture performances for the sequence learning control experiment}\label{tab:cfl-01-12-1-pareto}%
\begin{tabular*}{250pt}{@{\extracolsep{\fill}}lcc@{\extracolsep{\fill}}}
\toprule
Architecture & MSE loss & Number of blocks \\
\midrule
rdm32\_45    & \textbf{0.00115}   & 18     \\
rdm32\_32    & 0.00272   & 17     \\
rdm32\_26    & 0.03956   & 16     \\
rdm72\_41    & 0.04896   & 14     \\
rdm54\_29    & 0.06305   & 13     \\
BASIC\_20    & 0.07508   & 12     \\
BASIC\_0     & 0.13433   & \textbf{10}  \\
\botrule
\end{tabular*}
\end{minipage}
\end{center}
\end{table}

\subsection{Sentiment Analysis}
\label{sec:experiments:sentiment}
This section discusses the experimental results obtained after implementing the MOE/RNAS algorithm to search for and optimise RNN architectures for a sentiment analysis task based on the the ACL-IMBD \cite{Maas2011} dataset. The models were implemented with an embedding layer dimension of 1000 and a hidden layer dimension of 65, and a batch size of 50 was used during model training. Training and testing models on this sentiment analysis task consist of presenting the model with a sentence, and the model is then expected to predict whether the input sentence has a positive or a negative sentiment. Therefore, the model accuracy objective considered for this dataset is based on the number of predictions that were correct out of all the sentences from the testing data, which is then represented as a percentage value.

An LSTM model implemented for this task was able to achieve an accuracy of 83.10\% after being trained for only 5 epochs. The GRU architecture achieved an accuracy of 83.26\%, whereas the basic RNN architecture achieved an accuracy of 80\%.

The MOE/RNAS algorithm was implemented to search for and optimise RNN architectures for this task over a maximum of 15 generations and a population size of 100. The initial population included the basic RNN architecture and 99 RNN architectures were uniformly sampled from the search space.

The total search cost for this task was 20 GPU days, which is relatively high compared to the search costs observed from the previous tasks. Despite the low number of epochs that the models were trained for, the average training time was around 15 minutes per model, which played a significant role in the high search cost of this task.

The MOE/RNAS algorithm was able to successfully optimise the RNN architectures to outperform the LSTM and GRU architectures, with the best performing architecture found achieving an accuracy of 85.22\%. Furthermore, the best performing architecture was also the architecture with the lowest number of blocks amongst the Pareto-optimal architectures, which can be seen in Table~\ref{tab:sentiment_task_1}.

\begin{table}[ht]
\begin{center}
\begin{minipage}{250pt}
\caption{Pareto-optimal architecture performances for the sentiment analysis task}\label{tab:sentiment_task_1}%
\begin{tabular*}{250pt}{@{\extracolsep{\fill}}lcc@{\extracolsep{\fill}}}
\toprule
Architecture & Accuracy & Number of blocks \\
\midrule
rdm43\_40    & \textbf{85.22}  & \textbf{18}     \\
rdm69\_43    & 84.96           & 20     \\
rdm72\_28    & 84.48           & 21     \\
rdm48\_4     & 84.34           & 22     \\
rdm69\_79    & 83.92           & 23     \\
\botrule
\end{tabular*}
\end{minipage}
\end{center}
\end{table}

The experiment was then repeated using the same configuration, and after 15 generations and a search cost of 20 GPU days, better performing architectures were found compared to the first run of this experiment in terms of the model accuracy objective. The Pareto-optimal architectures found during this experimental run can be seen in Table~\ref{tab:sentiment_task_2}.

\begin{table}[ht]
\begin{center}
\begin{minipage}{250pt}
\caption{Pareto-optimal architecture performances for the second run of the sentiment analysis task}\label{tab:sentiment_task_2}%
\begin{tabular*}{250pt}{@{\extracolsep{\fill}}lcc@{\extracolsep{\fill}}}
\toprule
Architecture & Accuracy & Number of blocks \\
\midrule
rdm96\_26   & \textbf{86.64}  & \textbf{18}     \\
rdm5\_25    & 86.46           & 19     \\
rdm96\_62   & 86.42           & 22     \\
rdm5\_104   & 86.28           & 24     \\
rdm96\_63   & 83.90           & 26     \\
\botrule
\end{tabular*}
\end{minipage}
\end{center}
\end{table}

\subsection{Results Discussion}
\label{sec:experiments:discussion}
He \textit{et al.} \cite{He2021} postulated that current RNN NAS methods have yet to find novel RNN architectures that outperform state-of-the-art manually designed RNN architectures, specifically within the NLP domain. According to He \textit{et al.} \cite{He2021}, the best performing RNN architecture found by existing RNN NAS publications is the RNN cell discovered by the DARTS NAS method \cite{Liu2018}, which achieved a test perplexity of 56.1 on the Penn Treebank dataset.

The best performing architecture found by MOE/RNAS achieved a test perplexity of 92.7 on the Penn Treebank dataset with a total of 13.8M trainable parameters. In comparison, the DARTS cell has 33M trainable parameters \cite{Liu2018}. Therefore, the architecture found by the MOE/RNAS algorithm has a much lower computational resource demand, but at the cost of reduced model accuracy.

From the experiments performed on the sequence learning task, it was observed that the number of consecutive network transformations considered during network morphism has a significant contribution towards the MOE/RNAS algorithm's ability to optimise multiple RNN architecture objectives. When the maximum number of consecutive network transformations considered are too high, the RNN architectures optimised by the MOE/RNAS algorithm do not outperform those randomly created during the initialisation of the initial population.

Although the search for the sequence learning task was terminated after 15 generations, the total search cost was significantly lower compared to the more than 8 GPU days search cost of the experiments that used the word-level NLP task's dataset. This lower search cost compared to the search cost of the word-level NLP experiments was due to a significantly smaller training dataset. Additionally, since only 25 offspring architectures were created per generation, fewer models had to be trained per generation. It was observed during the control experiment for the sequence learning task that the 15 generation search cost increased to 1.78 GPU days when 100 offspring architectures were created per generation.

However, the 8 GPU days search cost of the experiments that used the word-level NLP task's dataset is significant, since the 8 GPU days search cost shows that the MOE/RNAS algorithm has a higher overall computational demand compared to the DARTS NAS method \cite{Liu2018}. This observation was further confirmed with the sentiment analysis experiments from Section~\ref{sec:experiments:sentiment}. The higher search cost of the MOE/RNAS algorithm is attributed to the training of the RNNs at each generation, despite the implementation of network morphism and early stopping to make the MOE/RNAS algorithm more efficient. Training and testing 100 RNN models at each generation is expected to have a high computational demand, and the methods implemented to make model accuracy evaluation more efficient were limited such that multi-objective RNN architecture evolution can be studied in more detail instead.

\section{Conclusion}
\label{sec:conclusion}
In this paper, we proposed the MOE/RNAS algorithm as a multi-objective EA-based NAS method for automated RNN architecture search, which was specifically developed to optimise both the model accuracy objective along with the RNN architecture complexity objective. The MOE/RNAS algorithm relies on methods such as network morphism and early stopping to make the generational RNN architecture evolution more efficient.

The experimental results obtained showed that the MOE/RNAS algorithm was able to automatically construct novel RNN architectures that can learn from the provided dataset. Additionally, it was observed that the MOE/RNAS algorithm is fully capable of optimising RNN architecture complexity-related objectives, and when a reasonable trade-off is accepted between model accuracy and the computational resources demanded by the model, the MOE/RNAS algorithm can evolve computationally efficient RNN architectures that achieve reasonably good model accuracy.

The MOE/RNAS algorithm was unable to find and evolve a novel RNN architecture that outperformed the current state-of-the-art RNN architectures in terms of test perplexity on the Penn Treebank dataset. However, RNN architectures were discovered that achieved comparable perplexity, but with significantly lower computational cost. Furthermore, the MOE/RNAS algorithm was able to find and evolve Pareto-optimal RNN architectures that dominated the manually designed RNN architectures, such as the LSTM.

It was observed that the approximate RNN morphism is sensitive to the maximum number of consecutive network transformations allowed during offspring generation. Lower numbers of consecutive network transformations result in a more consistent generational optimisation of the multiple objectives considered.

Overall, the MOE/RNAS algorithm is a good candidate for real-world machine learning applications where the model computational resource demand is of concern. Additionally, the MOE/RNAS method will be beneficial to use cases where the knowledge of existing pretrained models can be leveraged to search for models with reduced computational resource demand while maintaining an acceptable model accuracy objective.

An obvious avenue for future work would be to enhance the network morphism approach of the MOE/RNAS algorithm, which could include an RL agent to consider the impact of previous network transformations on the resulting RNN architecture fitness. Furthermore, performance prediction techniques, such as the density estimation technique implemented in \cite{Elsken2018}, can be incorporated to improve the overall search cost of the MOE/RNAS algorithm.

\section{Acknowledgements}
\label{sec:acknowledgements}
The authors would like to thank the Centre for High Performance Computing (CHPC) (\url{https://www.chpc.ac.za/}) for the use of their cluster to obtain the data for this study.

\section*{Declarations}
The authors have no relevant financial or non-financial interests to disclose.

\section*{Data Availability}
\begin{enumerate}
    \item The Penn Treebank dataset used for the word-level NLP task in Section~\ref{sec:experiments:ptb} is available for download at: \url{https://github.com/reinn-cs/rnn-nas/tree/master/example_datasets/ptb/data}.
    
    \item The dataset used for the sequence learning task in Section~\ref{sec:experiments:cfl} is artificially generated as described in the relevant section. The source code for the generation of the dataset is included in the source code repository of the MOE/RNAS algorithm implementation, which is available at: \url{https://github.com/reinn-cs/rnn-nas}.
    
    \item The data used for the analysis of the MOE/RNAS algorithm was based on the experimental results obtained after implementing the MOE/RNAS algorithm to search for and optimise RNN architectures for the respective datasets. The source code for the MOE/RNAS algorithm implementation is available at: \url{https://github.com/reinn-cs/rnn-nas}. 
\end{enumerate}



\begin{thebibliography}{2}

\bibitem{Mandic2001} D.P. Mandic, J.A. Chambers, Recurrent Neural Networks for Prediction, Wiley Series in Adaptive and Learning Systems for Signal Processing, Communications, and Control, vol. 4 (John Wiley \& Sons, Ltd, Chichester, UK, 2001), p. 297. \url{https://doi.org/10.1002/047084535X}

\bibitem{Medsker2013} L.R. Medsker, L.C. Jain, Recurrent Neural Networks: Design and Applications, 1st edn. (CRC Press, 2001)

\bibitem{Pascanu2014} R. Pascanu, C. Gulcehre, K. Cho, Y. Bengio, How to construct deep recurrent neural networks. In Proceedings of the Second International Conference on Learning Representations (ICLR 2014) (2014)

\bibitem{gpt2} I.S. Alec Radford, Jeffrey Wu, Rewon Child, David Luan, Dario Amodei, Language Models are Unsupervised Multitask Learners. OpenAI Blog 1(May), 1–7 (2020). URL \url{https://github.com/codelucas/newspaper}

\bibitem{Merity2018} S. Merity, N.S. Keskar, R. Socher, Regularizing and Optimizing LSTM Language Models. In Proceedings of the 6th International Conference on Learning Representations, ICLR 2018, Vancouver, BC, Canada, April 30 - May 3, 2018, Conference Track Proceedings. Retrieved from \url{https://openreview.net/forum?id=SyyGPP0TZ}

\bibitem{Suzgun2018} M. Suzgun, Y. Belinkov, S.M. Shieber, On evaluating the generalization of LSTM models in formal languages. In Proceedings of the Society for Computation in Linguistics, vol. 2 (2019), pp. 277–286. \url{https://doi.org/10.7275/s02b-4d91}

\bibitem{Wang2020} C. Wang, H. Wang, G. Feng, F. Geng, Multi-Objective Neural Architecture Search Based on Diverse Structures and Adaptive Recommendation. arXiv preprint arXiv:2007.02749 (2020). arXiv:2007.02749

\bibitem{Yang2019} Z. Yang, Y. Wang, X. Chen, B. Shi, C. Xu, C. Xu, Q. Tian, C. Xu, Cars: Continuous evolution for efficient neural architecture search. In Proceedings of the IEEE/CVF Conference on Computer Vision and Pattern Recognition (2019), pp. 1829–1838

\bibitem{Zoph2017} B. Zoph, Q.V. Le, Neural Architecture Search with Reinforcement Learning. In Proceedings of the 5th International Conference on Learning Representations, ICLR 2017, Toulon, France, April 24-26, 2017, Conference Track Proceedings. Retrieved from \url{https://openreview.net/forum?id=r1Ue8Hcxg}

\bibitem{Elsken2018} T. Elsken, J.H. Metzen, F. Hutter, Efficient Multi-Objective Neural Architecture Search via Lamarckian Evolution. In Proceedings of the 7th International Conference on Learning Representations, ICLR 2019, New Orleans, LA, USA, May 6-9, 2019. Retrieved from \url{https://openreview.net/forum?id=ByME42AqK7}

\bibitem{Liu2020} Y. Liu, Y. Sun, B. Xue, M. Zhang, G.G. Yen, K.C. Tan, A Survey on Evolutionary Neural Architecture Search. IEEE Transactions on Neural Networks and Learning Systems pp. 1–21 (2021). \url{https://doi.org/10.1109/TNNLS.2021.3100554}

\bibitem{Wistuba2019} M. Wistuba, A. Rawat, T. Pedapati, A Survey on Neural Architecture Search. arXiv preprint arXiv:1905.01392 (2019). arXiv:1905.01392

\bibitem{Lu2020} Z. Lu, I. Whalen, Y. Dhebar, K. Deb, E. Goodman, W. Banzhaf, V.N. Boddeti, NSGA-Net: Neural architecture search using multi-objective genetic algorithm. In Proceedings of the Twenty-Ninth International Joint Conference on Artificial Intelligence (International Joint Conferences on Artificial Intelligence Organization, 2020), pp. 4750–4754. \url{https://doi.org/10.24963/ijcai.2020/659}

\bibitem{Chen2020} Z. Chen, F. Zhou, G. Trimponias, Z. Li, Multi-objective Neural Architecture Search via Non-stationary Policy Gradient. arXiv preprint arXiv:2001.08437 (2020). arXiv:2001.08437

\bibitem{Klyuchnikov2020} N. Klyuchnikov, I. Trofimov, E. Artemova, M. Salnikov, M. Fedorov, A. Filippov, E. Burnaev, NAS-Bench-NLP: Neural Architecture Search Benchmark for Natural Language Processing. IEEE Access 10, 45,736–45,747 (2022). \url{https://doi.org/10.1109/ACCESS.2022.3169897}

\bibitem{Hu2020} S. Hu, R. Cheng, C. He, Z. Lu, Multi-objective Neural Architecture Search with Almost No Training. In Proceedings of the 11th International Conference on Evolutionary Multi-Criterion Optimization (Springer, Cham, 2021), (pp. 492–503). \url{https://doi.org/10.1007/978-3-030-72062-9\_39}

\bibitem{Chu2019} X. Chu, B. Zhang, R. Xu, H. Ma, Multi-objective reinforced evolution in mobile neural architecture search. In Proceedings of the European Conference on Computer Vision (Springer, Cham, 2020), pp. 99–113. \url{https://doi.org/10.1007/978-3-030-66823-5\_6}

\bibitem{Smith2015} J.E. Smith, A. Eiben, Introduction to Evolutionary Computing, 2nd edn. (Springer Publishing Company Inc., 2015)

\bibitem{Rudolph1998} G. Rudolph, On a multi-objective evolutionary algorithm and its convergence to the Pareto set. In Proceedings of the IEEE Conference on Evolutionary Computation, ICEC (IEEE, 1998), pp. 511–516. \url{https://doi.org/10.1109/icec.1998.700081}

\bibitem{Zhou2011} A. Zhou, B.Y. Qu, H. Li, S.Z. Zhao, P.N. Suganthan, Q. Zhangd, Multiobjective evolutionary algorithms: A survey of the state of the art. Swarm and Evolutionary Computation 1(1), 32–49 (2011). \url{https://doi.org/10.1016/j.swevo.2011.03.001}

\bibitem{Zitzler2000} E. Zitzler, K. Deb, L. Thiele, Comparison of Multiobjective Evolutionary Algorithms: Empirical Results. Evolutionary Computation 8(2), 173–195 (2000). \url{https://doi.org/10.1162/106365600568202}

\bibitem{CoelloCoello2002} C.A. Coello Coello, D.A. Van Veldhuizen, G.B. Lamont, Evolutionary Algorithms for Solving Multi-Objective Problems. Genetic and Evolutionary Computation Series (Springer US, Boston, MA, 2007). \url{https://doi.org/10.1007/978-0-387-36797-2}

\bibitem{Liu2018} H. Liu, K. Simonyan, Y. Yang, Darts: Differentiable architecture search. arXiv preprint arXiv:1806.09055 (2018). arXiv:1806.09055

\bibitem{Pham2018} H. Pham, M.Y. Guan, B. Zoph, Q.V. Le, J. Dean, Efficient Neural Architecture Search via Parameters Sharing. In Proceedings of the 35th International Conference on Machine Learning, vol. 80 (PMLR, 2018), pp. 4095–4104

\bibitem{Li2019} L. Li, A. Talwalkar, Random Search and Reproducibility for Neural Architecture Search. In Proceedings of the Thirty-Fifth Conference on Uncertainty in Artificial Intelligence, UAI 2019, Tel Aviv, Israel, July 22-25, 2019 (pp. 367–377). Retrieved from \url{http://proceedings.mlr.press/v115/li20c.html}

\bibitem{Yang2018} Z. Yang, Z. Dai, R. Salakhutdinov, W.W. Cohen, Breaking the Softmax Bottleneck: A High-Rank RNN Language Model. arXiv preprint arXiv:1711.03953 (2017). arXiv:1711.03953

\bibitem{Bayer2009} J. Bayer, D. Wierstra, J. Togelius, J. Schmidhuber, Evolving Memory Cell Structures for Sequence Learning. Artificial Neural Networks - ICANN 2009 pp. 755–764 (2009). \url{https://doi.org/10.1007/978-3-642-04277-5\_76}

\bibitem{Wei2016a} T. Wei, C. Wang, Y. Rui, C.W. Chen, Network morphism. In Proceedings of The 33rd International Conference on Machine Learning, vol. 48 (PMLR, New York, New York, USA, 2016), pp. 564–572

\bibitem{Cai2018cc} H. Cai, T. Chen, W. Zhang, Y. Yu, J. Wang, Efficient architecture search by network transformation. Proceedings of the AAAI Conference on Artificial Intelligence 32(1), 2787–2794 (2018). arXiv:1707.04873

\bibitem{Chen2016} G. Chen, A Gentle Tutorial of Recurrent Neural Network with Error Backpropagation. arXiv preprint arXiv:1610.02583 (2016). URL \url{http://arxiv.org/abs/1610.02583}. arXiv:1610.02583

\bibitem{Bengio1994} Y. Bengio, P. Simard, P. Frasconi, Learning long-term dependencies with gradient descent is difficult. IEEE Transactions on Neural Networks 5(2), 157–166 (1994). \url{https://doi.org/10.1109/72.279181}

\bibitem{Pascanu2013} R. Pascanu, T. Mikolov, Y. Bengio, On the difficulty of training recurrent neural networks. In Proceedings of the International conference on machine learning (PMLR, 2013), pp. 1310–1318

\bibitem{Hochreiter1997} S. Hochreiter, J. Schmidhuber, Long Short-Term Memory. Neural Computation 9(8), 1735–1780 (1997). \url{https://doi.org/10.1162/neco.1997.9.8.1735}

\bibitem{Kong2019} W. Kong, Z.Y. Dong, Y. Jia, D.J. Hill, Y. Xu, Y. Zhang, Short-Term Residential Load Forecasting Based on LSTM Recurrent Neural Network. IEEE Transactions on Smart Grid 10(1), 841–851 (2019). \url{https://doi.org/10.1109/TSG.2017.2753802}

\bibitem{Karpathy2015} A. Karpathy, J. Johnson, L. Fei-Fei, Visualizing and Understanding Recurrent Networks. arXiv preprint arXiv:1506.02078 (2015). arXiv:1506.02078

\bibitem{Chung2015} J. Chung, C. Gulcehre, K. Cho, Y. Bengio, Gated feedback recurrent neural networks. In Proceedings of the 32nd International Conference on Machine Learning, vol. 37 (PMLR, 2015), pp. 2067–2075

\bibitem{Jozefowicz2015} R. Jozefowicz, W. Zaremba, I. Sutskever, An Empirical Exploration of Recurrent Network Architectures. In Proceedings of the 32nd International Conference on Machine Learning, vol. 37 (JMLR.org, Lille, France, 2015), pp. 2332–2340

\bibitem{Cho2014} K. Cho, B. Van Merri¨enboer, C. Gulcehre, D. Bahdanau, F. Bougares, H. Schwenk, Y. Bengio, Learning phrase representations using RNN encoder-decoder for statistical machine translation. In Proceedings of the Empirical Methods in Natural Language Processing (ACL, 2014), pp.1724–1734. \url{https://doi.org/10.3115/v1/d14-1179}

\bibitem{Chung2014} J. Chung, C. Gulcehre, K. Cho, Y. Bengio, Empirical Evaluation of Gated Recurrent Neural Networks on Sequence Modeling. In Proceedings of the NIPS 2014 Workshop on Deep Learning, December 2014 (2014), pp. 1–9. URL \url{http://arxiv.org/abs/1412.3555}

\bibitem{engbr2007} A.P. Engelbrecht, Computational Intelligence: An Introduction, 2nd edn. (Wiley Publishing, 2007)

\bibitem{Ma2019} X. Ma, X. Li, Q. Zhang, K. Tang, Z. Liang, W. Xie, Z. Zhu, A Survey on Cooperative Co-Evolutionary Algorithms. IEEE Transactions on Evolutionary Computation 23(3), 421–441 (2019). \url{https://doi.org/10.1109/TEVC.2018.2868770}

\bibitem{Olague2016} A. Eiben, M. Schoenauer, Evolutionary computing. Information Processing Letters 82(1), 1–6 (2002). \url{https://doi.org/10.1007/978-3-662-43693-6\_3}

\bibitem{Zeng2003} SanYou Zeng, LiXin Ding, LiShan Kang, Yuping Chen, A new multiobjective evolutionary algorithm: OMOEA. Congress on Evolutionary Computation, 2003. CEC ’03 2, 898–905 (2003). \url{https://doi.org/10.1109/CEC.2003.1299762}. URL \url{http://ieeexplore.ieee.org/document/1299762/}

\bibitem{Deb2002} K. Deb, A. Pratap, S. Agarwal, T. Meyarivan, A fast and elitist multiobjective genetic algorithm: NSGA-II. IEEE Transactions on Evolutionary Computation 6(2), 182–197 (2002). \url{https://doi.org/10.1109/4235.996017}

\bibitem{Chu2018} X. Chu, X. Yu, Improved Crowding Distance for NSGA-II. arXiv preprint arXiv:1811.12667 (2018). arXiv:1811.12667

\bibitem{Lu2020a} Z. Lu, K. Deb, E. Goodman, W. Banzhaf, V.N. Boddeti, NSGANetV2: Evolutionary Multi-objective Surrogate-Assisted Neural Architecture Search. In Proceedings of the European Conference on Computer Vision (Springer, Cham, 2020), pp. 35–51. \url{https://doi.org/10.1007/978-3-030-58452-8\_3}

\bibitem{Lu2021} Z. Lu, I. Whalen, Y. Dhebar, K. Deb, E.D. Goodman, W. Banzhaf, V.N. Boddeti, Multiobjective Evolutionary Design of Deep Convolutional Neural Networks for Image Classification. IEEE Transactions on Evolutionary Computation 25(2), 277–291 (2021). \url{https://doi.org/10.1109/TEVC.2020.3024708}

\bibitem{Park2020} K.m. Park, D. Shin, Y. Yoo, Evolutionary Neural Architecture Search (NAS) Using Chromosome Non-Disjunction for Korean Grammaticality Tasks. Applied Sciences 10(10), 3457 (2020). \url{https://doi.org/10.3390/app10103457}

\bibitem{He2021} X. He, K. Zhao, X. Chu, AutoML: A survey of the state-of-the-art. Knowledge-Based Systems 212(Dl) (2021). \url{https://doi.org/10.1016/j.knosys.2020.106622}

\bibitem{White2020} C. White, W. Neiswanger, S. Nolen, Y. Savani, A Study on Encodings for Neural Architecture Search. In Proceedings of the Advances in Neural Information Processing Systems 33: Annual Conference on Neural Information Processing Systems 2020, NeurIPS 2020, December 6-12, 2020, virtual. Retrieved from \url{https://proceedings.neurips.cc/paper/2020/hash/ea4eb49329550caaa1d2044105223721-Abstract.html}

\bibitem{Mo2021} H. Mo, L.L. Custode, G. Iacca, Evolutionary neural architecture search for remaining useful life prediction. Applied Soft Computing 108 (2021), 107474. \url{https://doi.org/10.1016/j.asoc.2021.107474}


\bibitem{Angeline1994} P. Angeline, G. Saunders, J. Pollack, An evolutionary algorithm that constructs recurrent neural networks. IEEE Transactions on Neural Networks 5(1), 54–65 (1994). \url{https://doi.org/10.1109/72.265960}

\bibitem{Paszke2019} A. Paszke, S. Gross, F. Massa, A. Lerer, J. Bradbury, G. Chanan, T. Killeen, Z. Lin, N. Gimelshein, L. Antiga, A. Desmaison, A. K¨opf, E. Yang, Z. DeVito, M. Raison, A. Tejani, S. Chilamkurthy, B. Steiner, L. Fang, J. Bai, S. Chintala, PyTorch: An imperative style, highperformance deep learning library. Advances in Neural Information Processing Systems (2019). arXiv:1912.01703

\bibitem{Jozefowicz2016} R. Jozefowicz, O. Vinyals, M. Schuster, N. Shazeer, Y. Wu, Exploring the Limits of Language Modeling. arXiv preprint arXiv:1602.02410 (2016). arXiv:1602.02410

\bibitem{Jurafsky2014} D. Jurafsky, J. Martin, Speech and Language Processing (Prentice-Hall, Inc., 2008)

\bibitem{Maas2011} A. Maas, R. Daly, P. Pham, D. Huang, A. Ng, C. Potts, Learning word vectors for sentiment analysis. In Proceedings of the 49th annual meeting of the association for computational linguistics: Human language technologies (ACL, 2011), pp. 142-150.

\bibitem{Nugaliyadde2019} A. Nugaliyadde, F. Sohel, K. W. Wong and H. Xie, Language Modeling through Long-Term Memory Network. In Proceedings of the 2019 International Joint Conference on Neural Networks (IJCNN), Budapest, Hungary, 2019, pp. 1-6. \url{https://doi.org/10.1109/IJCNN.2019.8851909}

\end{thebibliography}
\end{document}